\pdfoutput=1
\pdfmapfile{+cjhebrew/cjhebrew.map}

\documentclass[11pt]{article}

\usepackage[]{EACL2023}

\usepackage{times}
\usepackage{latexsym}
\usepackage{cjhebrew}

\usepackage{placeins}
\usepackage{comment}
\usepackage[T1]{fontenc}
\usepackage{svg}

\usepackage{graphicx}
\usepackage{array}
\usepackage{longtable}
\usepackage{multirow}
\usepackage{titlesec}
\newcolumntype{?}{!{\vrule width 1.5pt}}
\newcolumntype{@}{!{\vrule width 3pt}}
\usepackage{makecell}
\usepackage{etoolbox}
\usepackage{caption}
\usepackage{lipsum}
\usepackage{enumitem,kantlipsum}
\usepackage{svg}
\usepackage{booktabs}

\usepackage[utf8]{inputenc}

\usepackage{microtype}

\usepackage{inconsolata}

\makeatletter
\newcommand{\thickhline}{%
    \noalign {\ifnum 0=`}\fi \hrule height 2pt
    \futurelet \reserved@a \@xhline
}
\newcolumntype{"}{@{\hskip\tabcolsep\vrule width 2pt\hskip\tabcolsep}}
\makeatother

\usepackage[T1]{fontenc}

\title{Do Pretrained Contextual Language Models Distinguish between \\ Hebrew Homograph Analyses?}

\author{Avi Shmidman\textsuperscript{1,2,†}, Cheyn Shmuel Shmidman\textsuperscript{2,‡}, \\ \textbf{Dan Bareket\textsuperscript{1,‡}, Moshe Koppel\textsuperscript{1,2,‡}, Reut Tsarfaty\textsuperscript{1,†}} \\
\textsuperscript{1}Bar Ilan University / Ramat Gan, Israel \quad 
\textsuperscript{2}DICTA / Jerusalem, Israel \\
\texttt{\small \textsuperscript{†}\{avi.shmidman,reut.tsarfaty\}@biu.ac.il} \\
\texttt{\small \textsuperscript{‡}\{cheynshmuel,dbareket,moishk\}@gmail.com}}

\begin{document}
\maketitle
\begin{abstract}

Semitic morphologically-rich languages (MRLs) are characterized by extreme word ambiguity. Because most vowels are omitted in standard texts, many  of the words are homographs with multiple possible analyses, each with a different pronunciation and different morphosyntactic properties. This ambiguity goes {\em beyond} word-sense disambiguation (WSD), and may include token segmentation into multiple word units. Previous research on MRLs claimed that standardly trained pre-trained language models (PLMs) based on word-pieces may not sufficiently capture the internal structure of such tokens in order to distinguish between these  analyses.
Taking Hebrew as a case study, we investigate the extent to which Hebrew homographs can be disambiguated and analyzed using PLMs. We evaluate all existing models for contextualized Hebrew embeddings on a novel Hebrew homograph challenge sets that we deliver. Our empirical results demonstrate that contemporary Hebrew contextualized embeddings outperform non-contextualized embeddings; and that they are most effective for disambiguating segmentation and morphosyntactic features, less so regarding pure word-sense disambiguation. We show that these embeddings are more effective when the number of word-piece splits is limited, and they are more effective for 2-way and 3-way ambiguities than for 4-way ambiguity. We show that the embeddings are equally effective for homographs of both balanced and skewed distributions, whether calculated as masked or unmasked tokens. Finally, we show that these embeddings are as effective for homograph disambiguation with extensive supervised training as with a few-shot setup.
\end{abstract}

\section{Introduction}
Semitic morphologically-rich languages (MRLs) such as Arabic, Hebrew, and Aramaic are characterized by extreme ambiguity at the word level \cite{Wintner2014MorphologicalPO,tsarfaty-etal-2020-spmrl}. In a standard text, many (and often most) of the words will be homographs with multiple possible analyses. The high ambiguity derives from several factors. First, prepositions, conjunctions, accusative pronouns, and possessive pronouns are often seamlessly affixed to words. 
Next, vowels are generally omitted in written texts. Finally, proper nouns are not differentiated from common nouns (no capital letters). 

The task of distinguishing between Hebrew homograph analyses is related to the general task of Word Sense Disambiguation (WSD) \cite{AgirreAndEdmonds2006a, navigli2009}, yet it is more challenging. In the standard case of WSD, a single orthographic form is associated with a single word that can be analyzed in terms of two or more senses; also, the analyses are generally pronounced identically, and often have the same morphosyntactic properties (e.g bank of a river vs. savings bank). In contrast, in Semitic languages, the need for disambiguation often goes beyond a determination of sense. Hebrew word ambiguities can be divided into three primary categories (Table \ref{SampleHomographs}): 1. Segmentation ambiguities, in which a given orthographic form may (or may not) be 
segmented into {multiple} word units each bearing its own role (POS tag) in the sentence.
2. Morphosyntactic ambiguities, in which the segmentation of the form is not ambiguous, but multiple analyses of the word reflect different morphosyntactic properties of each word unit(s). 3. Sense ambiguities (the aforementioned standard case of WSD), in which the analyses of the unit(s) do not differ in their morphosyntactic properties, but rather in their {\em sense}. One orthographic form may exhibit multiple types of ambiguity simultaneously.

Pretrained contextualized language models with standard word-piece tokenization mechanisms have been shown to excel at WSD in English and other Indo-European languages \cite{yaghoobzadeh-etal-2019-probing}. However, for Hebrew and other semitic languages it has been argued that such models would not sufficiently capture the structure of MRLs in order to distinguish between internally-complex homograph analyses  \cite{klein-tsarfaty-2020-getting,tsarfaty-etal-2020-spmrl}. In this work, we take Modern Hebrew, a Semitic language with rich and highly ambiguous morphology, as a case study, and we investigate the extent to which homographs can be disambiguated by contextualized embeddings, regarding all three levels of ambiguity. Regarding Arabic --- a sister language to Hebrew --- a wide survey of WSD methods is presented by \newcite{10.1145/3510451}. They raise the possibility of utilizing pretrained contextualized embeddings, yet leave its evaluation to future work.\footnote{Additional studies in Arabic WSD include \newcite{inproceedingsArabic}, \newcite{merhbene-etal-2013-semi}, and \newcite{shah-etal-2010-new}.}

\begin{table}[t]
\resizebox{\columnwidth}{!}{%
\begin{tabular}{|c|c"c|c|}
\hline
Type & Form & Word (translation) & Morphology \\ \thickhline
\multirow{4}{*}{Segmentation}
& \<hqph>
& \<q*ApEh>+\<ha> (the+coffee) & DET + Noun [M,S,abs]\\ \cline{3-4}
& & \<haq*ApAh> (credit) & Noun [F,S,abs]\\ \cline{2-4}
\noalign{\vskip-2\tabcolsep \vskip-3\arrayrulewidth \vskip\doublerulesep}
\\ \cline{2-4}
& \</s'P>
& \<'aP>+\<+sE>  (for+even) & Sconj + Cconj\\ \cline{3-4}
& & \<+sA'aP> (he aspired) & Verb [M,S,3,PAST]\\ \thickhline
\multirow{4}{*}{Morph}
& \<'lymwt>
& \<'al*iymwot> (violent) & Adj [F,P,abs]\\ \cline{3-4}
& & \<'al*iymw*t> (violence) & Noun [F,S,abs/cons]\\ \cline{2-4}
\noalign{\vskip-2\tabcolsep \vskip-3\arrayrulewidth \vskip\doublerulesep}
\\ \cline{2-4}
& \<hryM>
& \<heriyM> (he lifted) & Verb [M,S,3,PAST]\\ \cline{3-4}
& & \<hAriyM> (mountains) & Noun [M,P,abs]\\ \thickhline
\multirow{4}{*}{Semantic}
& \<hzmr>
& \<z*EmEr>+\<ha> (the+song) & DET + Noun [M,S,abs]\\ \cline{3-4}
& & \<z*am*Ar>+\<ha> (the+singer) & DET + Noun [M,S,abs]\\ \cline{2-4}
\noalign{\vskip-2\tabcolsep \vskip-3\arrayrulewidth \vskip\doublerulesep}
\\ \cline{2-4}
& \<hswpr>
& \<s*woper>+\<ha> (the+author) & DET + Noun [M,S,abs]\\ \cline{3-4}
& & \<s*w*p*Er>+\<ha> (the+market) & DET + Noun [M,S,abs]\\ \hline
\end{tabular}}
\caption{Examples of Hebrew ambiguity types}
\label{SampleHomographs}
\end{table}

Hebrew is a particularly challenging language on which to perform a homograph disambiguation due to the limited available corpora. First of all, currently existing Hebrew treebanks are severely limited in size, such that most of the words in the language are not amply represented. Furthermore, even regarding common Hebrew words, this corpus is problematic, because the nature of language is such that many homographs are skewed in their distribution; thus, even if the primary analysis is sufficiently represented within a tagged corpus, the secondary analysis will often be hopelessly underrepresented. 
For instance, one common Hebrew homograph is \<mhM> (\textit{mhm}), which can be analyzed as a preposition with pronominal suffix, or as an interrogative. 
The ratio of these two analyses in naturally-occurring Hebrew text is over 50:1; thus, occurrences of the secondary analysis within existing tagged corpora are  insufficient.

In theory, these homograph ambiguities could be addressed using POS tagging systems. For instance, \newcite{habash-rambow-2005-arabic} consider the use of a morphological tagging system to solve WSD in Arabic. A number of Hebrew POS tagging systems have been published as well \cite{yona-wintner-2005-finite, 387c824222424d8b853fc3427dea7867, shacham-wintner-2007-morphological}. The current SOTA for Hebrew POS tagging is the YAP morpho-syntactic parser \cite{ tsarfaty2019whats}. However, as we have shown in a previous study \cite[p. 3318, table 2]{shmidman-etal-2020-novel}, although YAP produces high accuracy overall on normal Hebrew text, its scores drop drastically regarding homographs of skewed distribution.

For analogous cases of skewed distribution in other languages, researchers have proposed the creation of dedicated challenge sets, containing hard-to-classify sentences not easily found in naturally-occurring text \cite{gardner2020evaluating, elkahky-etal-2018-challenge}.
In the aforementioned previous study, we produced 22 such challenge sets for Hebrew homographs, and demonstrated that a Bi-LSTM of non-contextualized  embeddings can obtain high accuracy on this task, establishing the current SOTA for Hebrew homograph disambiguation \cite{shmidman-etal-2020-novel}. In this paper, we extend the investigation by considering whether contextualized embeddings from pretrained language models (PLMs) can provide a more optimal solution. We consider all existing contextualized Hebrew PLMs: multilingual BERT ("mBERT") \cite{devlin2019bert};  HeBERT \cite{chriqui2021hebert}; and AlephBERT \cite{seker2021alephberta} (Table \ref{bertdata}). Moreover, we evaluate and verify these on a new dataset, substantially larger than all previous datasets for Hebrew homograph disambiguation.

Our experiments demonstrate that
contextualized PLMs pre-trained on sufficiently large unlabeled data and vocabulary size are excellent at disambiguating the word-internal structures of homographs, yet  face some challenge with pure sense disambiguation. We  show the efficacy of these models in cases of homographs with skewed distribution, and in  a few-shot setup. Finally, we establish  new state-of-the-art results on the challenging task of homograph disambiguation for a morphologically-rich language printed without vowels, along with a novel benchmark for assessing the morphological reach of future PLMs in Hebrew.


\section{The Data}
The challenge sets for Hebrew homograph disambiguation from our previous study were limited in number (only 22 sets) and insufficiently representative regarding types of ambiguities; only one of the sets was a prefix-segmentation ambiguity. Further, they were limited to binary cases, where only two analyses exist.

\begin{table}[]
    \centering
    \scalebox{.75}{
    \begin{tabular}{|c|c|c|}
    \hline
        Model & Vocab & Corpus Size \\ & (Heb. tokens) & (Heb. sentences) \\ \hline
        mBERT & 2.5K & 6.3M\\ \hline
        HeBERT & 30K & 27.2M \\ \hline
        AlephBERT & 52K & 98.7M \\ \hline
    \end{tabular}}
    \caption{Comparison of available Hebrew BERT models}
    \label{bertdata}
\end{table}
In contrast, for this study we 
employed field experts to choose the most critical homographs in the language. The experts chose 75 homographs 
from a list of the 3600 most frequent words in the language, balancing frequency of word occurrence with practical need for its disambiguation.
All of the homographs occur with a minimum frequency of 27 words per million in naturally occurring Hebrew text. 
Our challenge sets include homographs with 2-5 possible analyses. Our sets contain a wide representation of segmentation ambiguities (15 in number), as well as 5 cases of purely semantic ambiguities. 
For each of the 75 homographs, we collect hundreds of naturally-occurring sentences attesting to each analysis. In almost all cases, we succeed in collecting 1000 sentences for the primary analysis, at least 500 sentences for the secondary analysis, and at least 250 for each additional analysis. The sentences were culled from newspapers, Wikipedia, literature, and social media. We employed a team of annotators who chose the relevant homograph analysis for each case.\footnote{The annotation process is detailed in Appendix \ref{section:appendixCuration}.}
All in all, our 75 challenge sets contain 150K tagged sentences.
The full list of homographs and analyses is provided in Appendix \ref{section:AppendixHomographTable}.\footnote{The dataset is downloadable at: \url{https://github.com/Dicta-Israel-Center-for-Text-Analysis/EACL_2023}}

\section{Experimental Setup}
To evaluate the ability of pre-trained language models (PLMs) to disambiguate the in-context analyses of morphologically rich and highly ambiguous homographs in Hebrew, we adopt a "word expert" approach, producing dedicated classifiers for each individual homograph \cite{zhao-etal-2020-quantifying}. 

We use two types of PLMs, contextualized and non-contextualized. For the non-contextualized case, we replicate our previous method detailed in  \newcite{shmidman-etal-2020-novel}. For each training example, we use a BiLSTM on top of the word2vec embeddings of all of the words in the sentence (other than the homograph itself) to produce an encoding for disambiguation.\footnote{We also tested fastText, but results were inferior.} An MLP is trained to predict the correct homograph analysis based on this encoding.\footnote{For implementation details, see Appendix \ref{section:appendixImplementation}.}
For the contextualized case, we run the sentence through a pretrained contextualized language model and  retrieve the 768-dimension embedding representing the homograph in question. An MLP is trained to predict the correct analysis based on the homographs embeddings alone. In the standard "unmasked" scenario, the sentence is fed into the model as is, including the homograph in question. In the "masked" scenario, the homograph is replaced with a [MASK] token.

We evaluate the performance of each given method on each given challenge set using 10-fold cross-validation. We calculate an F1 score for each homograph analysis, based upon the precision and recall scores micro-averaged across all folds. We then calculate the macro-average of the F1 scores for all possible analyses for a given homograph, and this is the score  reported in the  charts herein.

\section{Results and Analysis}

\paragraph{Standard (Unmasked) Scenario}


\begin{figure}[t]
\centering
  \includegraphics[width=\columnwidth]{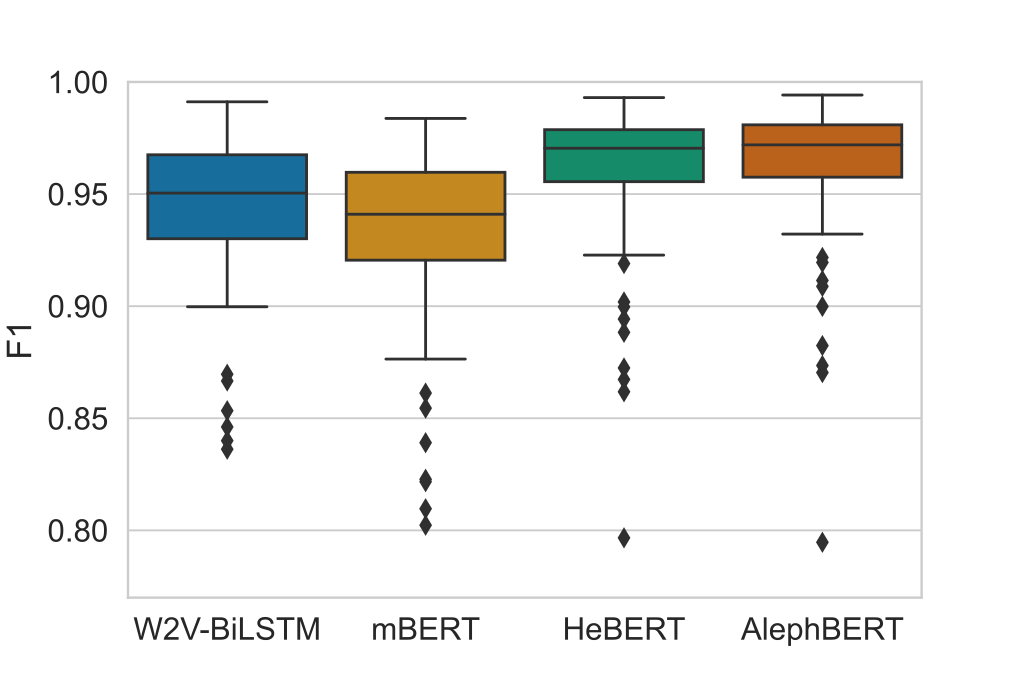}
  \caption{\label{fig:bert-vs-prev-sota} 
  Comparison of previous SOTA (w2v-based Bi-LSTM method) versus BERT-based approaches.}
\end{figure}


\begin{figure}[t]
\centering
  \includegraphics[width=\columnwidth]{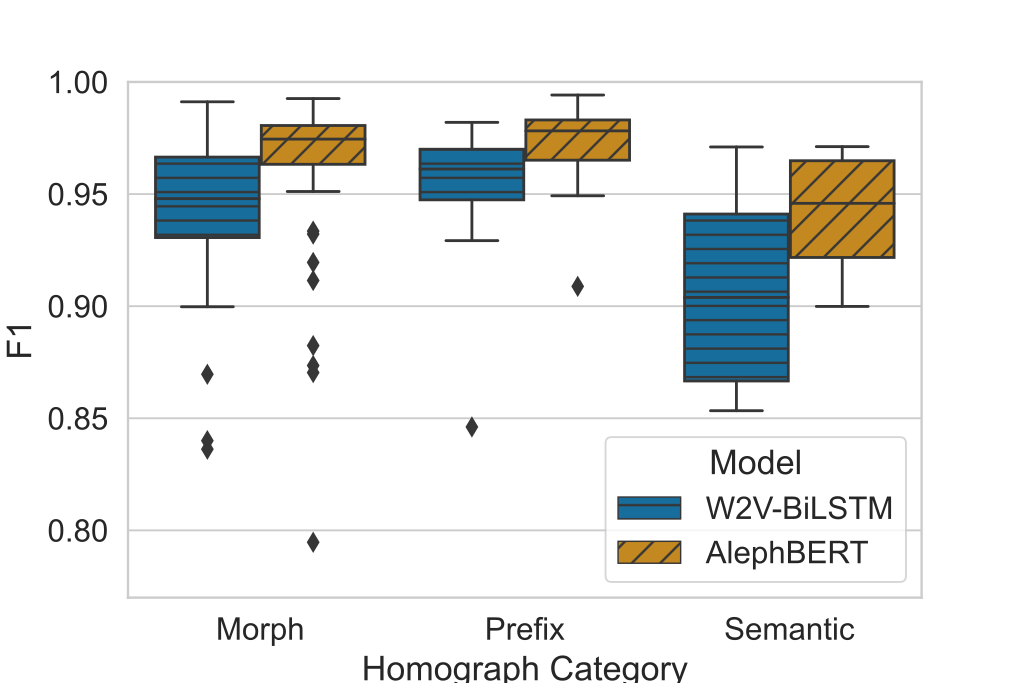}
  \caption{\label{fig:category} 
  Disambiguation accuracy across various categories of homograph ambiguity.}
\end{figure}


\begin{figure}[t]
\centering
  \includegraphics[width=\columnwidth]{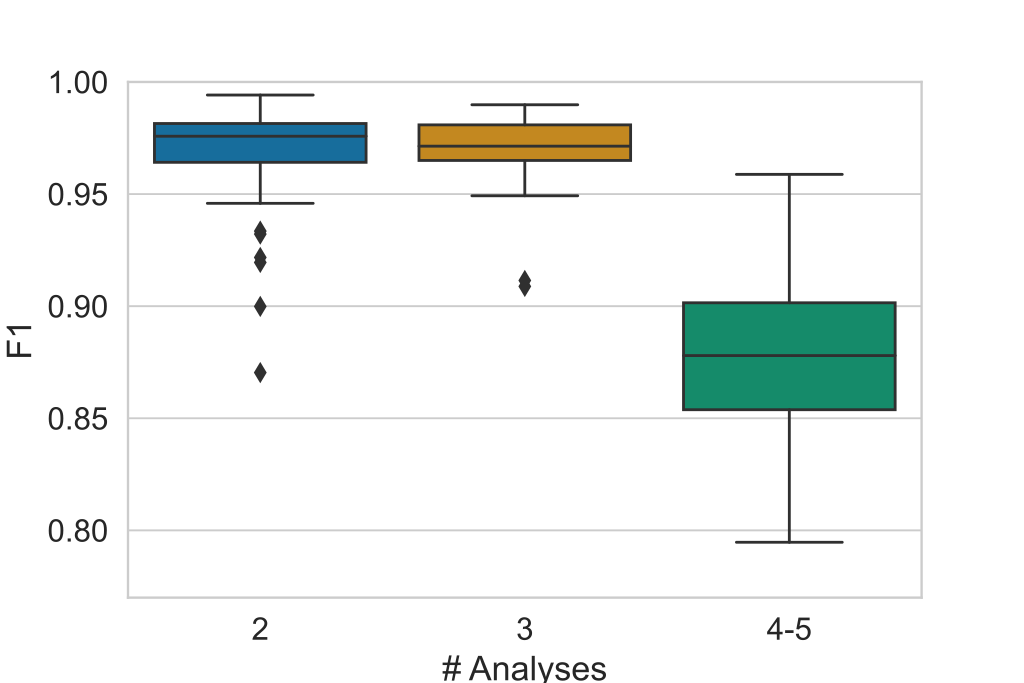}
  \caption{AlephBERT's disambiguation accuracy across homographs with differing counts of possible analyses.}
  \label{fig:num_analyses}
\end{figure}


\begin{figure}[t]
\centering
  \includegraphics[width=\columnwidth]{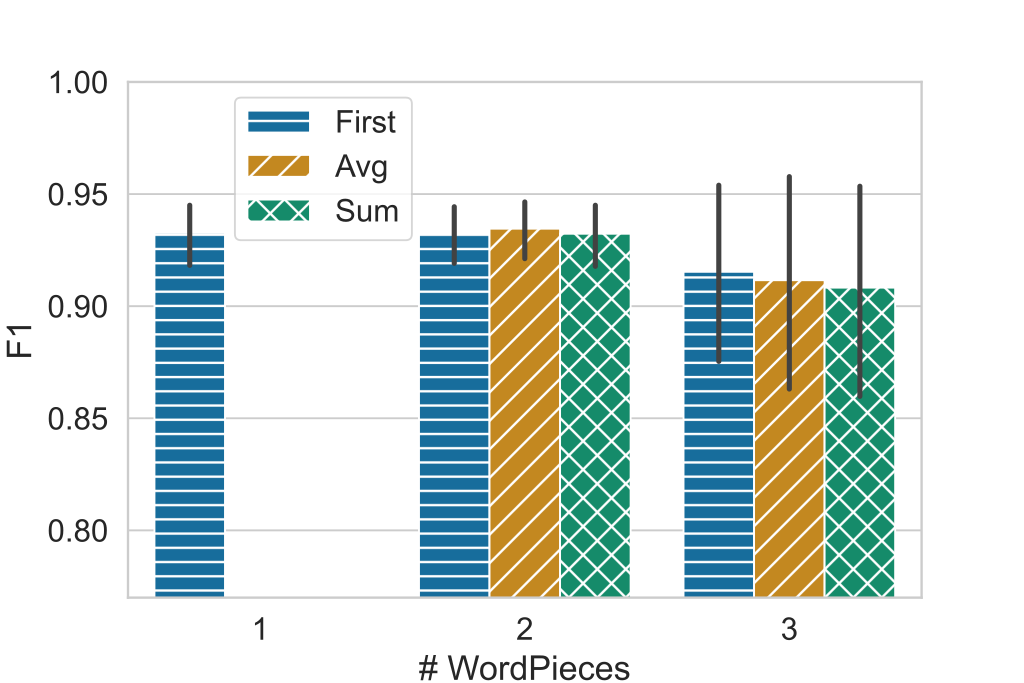}
  \caption{\label{fig:wordpiece} 
  Disambiguation accuracy across varying degrees of word piece splits within the target homograph, using mBERT.}
\end{figure}

Figure \ref{fig:bert-vs-prev-sota} presents the cumulative F1 score obtained by the models for all challenge sets. Our results show that  HeBERT and AlephBERT far outperform mBERT, with AlephBERT achieving the higher score. The poor performance of mBERT is likely due to its smaller pre-training data size and exceedingly lean Hebrew vocabulary (cf.\ Table \ref{bertdata}).
Furthermore, the HeBERT and AlephBERT models both substantially outperform the previous word2vec-based SOTA. It is thus apparent that {\em contextualized} language models do effectively capture Hebrew homograph distinctions, even those based on word-pieces, even for an MRL, and they do so more effectively than non-contextualized models.

Figure \ref{fig:category} demonstrates AlephBERT's performance on different ambiguity types. AlephBERT performs equally well on cases of segmentation ambiguity and  morphosyntactic ambiguity. In contrast, when it comes to ambiguities that are purely semantic, the scores are noticeably lower. This is in line with the findings of \newcite{ettinger-2020-bert}, who shows that BERT is stronger with syntax than semantics; \newcite{DBLP:journals/corr/abs-1901-05287} also notes BERT's strong syntactic abilities.
Interestingly, the same gap exists with the W2V-based method. 
Thus, both contextualized and non-contextualized embeddings struggle to differentiate between senses which are morphologically equivalent. 
Although such cases are only of minimal import when it comes to sentence parsing, they are critical for downstream tasks such as coreference resolution and relation extraction. 
It thus remains a desideratum to improve disambiguation of purely semantic Hebrew homographs.

The results in Figure \ref{fig:num_analyses} demonstrate that AlephBERT performs equally well on cases of binary homographs as on cases of three-way homograph classification. However, when faced with cases of 4-way or 5-way classification, accuracy declines.

\paragraph{The Effect of Word-Pieces}
Previous studies have hypothesized that word-pieces are not adequate for capturing complex morphosyntactic structures due to arbitrary (non-linguistic) word-splits. To probe into this we investigate the question, do such splits affect performance. Our 75 homographs are all treated as single tokens in HeBERT and AlephBERT. 
However, many of the homographs are broken up into word pieces in mBERT, due to its meager Hebrew vocabulary. We thus compare mBERT's results on words treated as single tokens versus those that are broken up into two or three pieces, which are aggregated using first, sum, or average of the  vectors. With regard to cases of split words, we train models using three separate methods: providing the MLP with only the embedding of the first word piece; with an average of the word piece embeddings; or with the sum of the embeddings. As shown in Figure \ref{fig:wordpiece}, the splitting of a homograph into three word-pieces appears to have a negative impact on the ability of the resulting embedding to differentiate between homograph analyses, for all aggregation methods. 

\paragraph{Masked Scenario}


\begin{figure}[t]
\centering
  \includegraphics[width=\columnwidth]{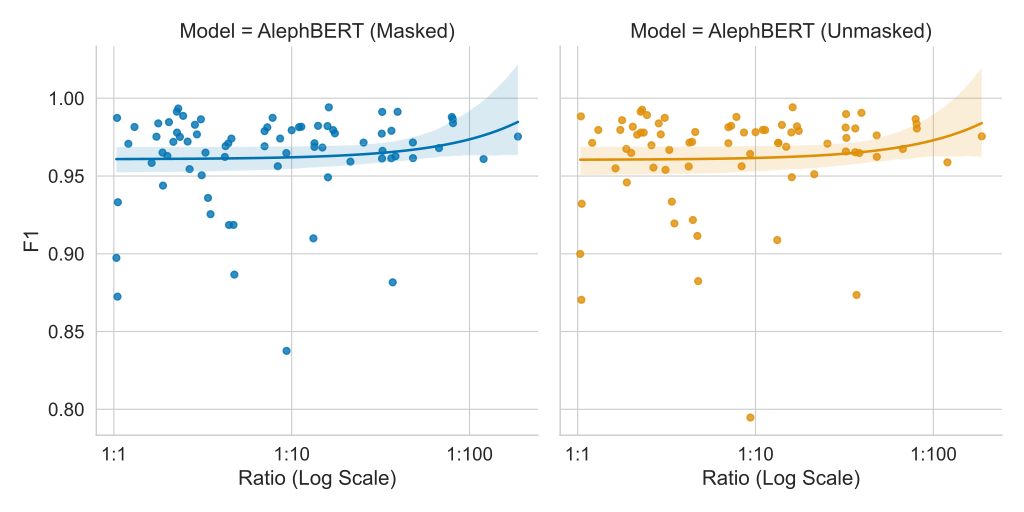}
  \caption{\label{fig:ratio22} 
  Comparison of unmasked vs. masked scenarios, across varying degrees of skewed homographs. The ratio axis indicates the relative distribution between the least and most frequent analyses of each plotted homograph within naturally-occurring Hebrew text.}
\end{figure}

We consider whether AlephBERT embeddings are more effective if we replace the homograph word with [MASK] when running the challenge set sentences through AlephBERT. The motivation behind this experiment is that, as explained above, 
many of the homographs are skewed in their natural proportion. In such cases, we worry whether AlephBERT might be disproportionately influenced by the skewed distribution; replacing the word with [MASK] would prevent the model from being influenced as such. As shown in figure \ref{fig:ratio22}, AlephBERT achieves high scores both with balanced homographs as well as with homographs of highly skewed distribution. Using a [MASK] token instead of the actual word does not generally improve performance, whether or not the homographs are of skewed proportion.

\begin{figure}[t]
\centering
  \includegraphics[width=\columnwidth]{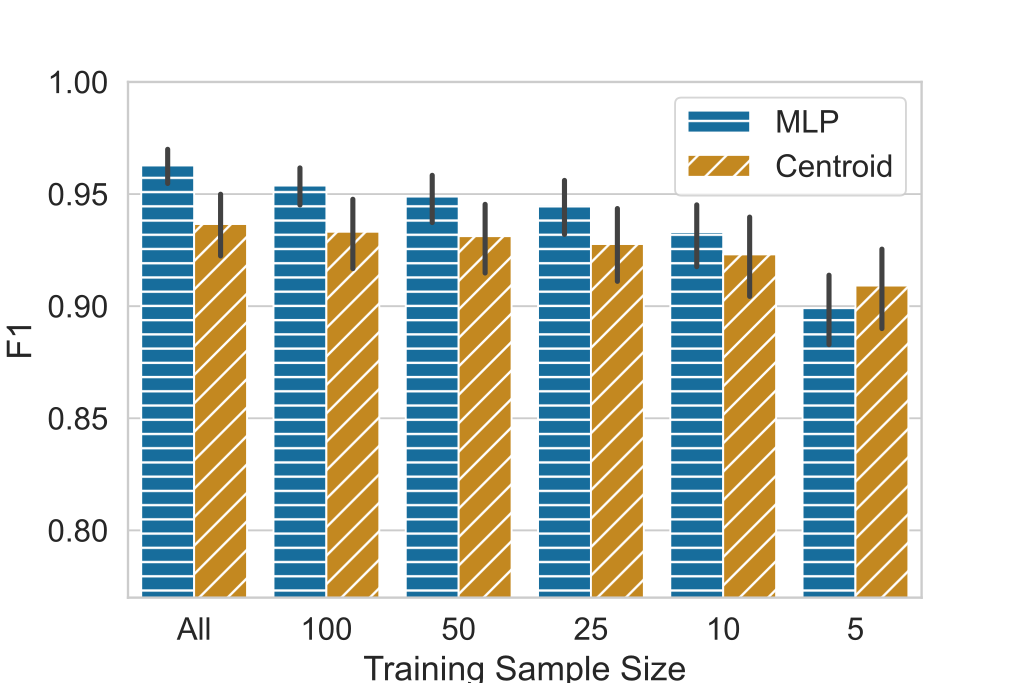}

  \caption{\label{fig:few-shot} 
  Use of AlephBERT embeddings to differentiate between homographs on a few-shot basis, contrasted with scores from the full 10-fold CV ("All").}
\end{figure}

\paragraph{Few-Shot Scenarios}
In our experiments thus far, the 10-fold cross-validation allows the MLP to leverage 90\% of the data in each fold (hundreds of sentences for each analysis)
in order to learn the difference between the analyses. We now consider whether the AlephBERT embeddings can suffice on a few-shot basis, where the training stage has access to only 100, 50, 25, 10 or even 5 examples of each analysis. In these cases, we train an MLP based only on these few samples, and we use the rest of the sentences for evaluation. Astoundingly, as demonstrated in Figure \ref{fig:few-shot}, the AlephBERT embeddings provide a highly accurate solution even on this few-shot basis. Even when training with only 5 examples of each homograph analysis, AlephBERT reaches an accuracy that is not far below the accuracy achieved when performing full 10-fold CV across hundreds of sentences of each analysis.

\paragraph{Probing Scenarios}
Finally, we probe the pretrained AlephBERT embeddings \cite{yaghoobzadeh-etal-2019-probing, tenney2019learn, klafka-ettinger-2020-spying, belinkov_probing2021} to see whether in and of themselves they reflect clusters which correspond to different homograph analyses.
We skip the MLP, and instead use the raw embeddings directly, classifying sentences based on their proximity to the centroid of the training samples for each homograph analysis.
As shown in the orange bars in Figure \ref{fig:few-shot}, this method generally does not perform as well as the MLP-based method; however, the degradation is limited to only a few percentage points, indicating that the raw embeddings are generally clustered in groups which indeed reflect the distinctions between the analyses. 


\section{Conclusion}
In this study we have utilized a wide-ranging collection of Hebrew homograph challenge sets in order to evaluate the extent to which raw contextualized embeddings can be leveraged to disambiguate Hebrew homographs. We found that contextualized embeddings can effectively disambiguate analyses of homographs, much more so than non-contextualized ones, regarding multiple types of ambiguity: segmentation, morphosyntactic and sense. Yet, efficacy on pure sense ambiguity is lower than on the other two types. Additionally, an increasing  number of splits, or an increasing number of different possible analyses of a token, each lower efficacy. 
Finally, we found that contextualized embeddings can function effectively for this purpose on a few-shot basis, with as little as 5 examples of each analysis. This indicates that with relatively modest effort, highly ambiguous homographs may be effectively treated.

\newpage

\section*{Limitations}
One of the major strengths of this paper is its new and comprehensive dataset for the training and benchmarking Hebrew homograph disambiguation. The dataset is uncomparable in size, quality and balance to all previous Hebrew homograph datasets; we have made every effort to be as inclusive as possible in the creation of the dataset, making sure to include data from a widely diverse set of genres. Nevertheless, a perennial challenge in corpus-based studies is that the lion's share of the available data tends to be authored by male writers. In order to offset this bias, we bolstered our corpus with a large corpus of texts specifically taken from blog sites devoted entirely to female bloggers. Even so, we cannot escape the fact that female writing and feminine conjugations are underrepresented in our dataset.

A further limitation derives from our filter for sentences with offensive language. We perceived early on that our human annotators were not comfortable tagging sentences with offensive language, and we therefore took steps to remove such sentences from our corpus. Nevertheless, this means that our resulting dataset is limited in that it does not properly reflect the use of offensive language in naturally-occurring Hebrew sentences. Furthermore, our resulting tests and reported scores may not accurately reflect the performance of our models when applied to sentences with offensive language.

\section*{Ethics Statement}
\paragraph{Creation of the Dataset}
As noted, our dataset contains over 150K sentences in all. Every sentence was reviewed and tagged by our team of human annotators, who chose the relevant homograph analysis for each instance of each of our 75 homographs. 
Our annotator team included members of diverse genders and sexual orientations. They were paid hourly wages with legal pay stubs. Their hourly wage was well above the minimum wage required by law. The entirety of the dataset will be made available for research purposes with the acceptance of this article, together with the tagging information.

\paragraph{Risks of the Research}
Ultimately, this data will enable end-users to automatically diacritize and parse large corpora of Hebrew text. For the most part, this will provide a beneficial contribution to the world: for the visually impaired, this technology will enable the development of more precise text-to-speech products; teachers will be able to provide children and second-language learners with accessible diacritized texts; and humanities and linguistics researchers can bolster their research with big-data analysis. However, there also is a risk of nefarious use, for instance, if an end user were to leverage these capabilities in order to produce anonymous texts or recordings containing threats to life, liberty, or happiness.


\section*{Acknowledgments}
The work of the third and last authors has been funded by a  grant from the Israeli Ministry of Science and Technology (MOST) grant no.\  3-17992. 

\bibliography{anthology,custom}
\bibliographystyle{acl_natbib}

\appendix

\onecolumn
\newpage

\def\hlinewd#1{%
\noalign{\ifnum0=`}\fi\hrule \@height #1 %
\futurelet\reserved@a\@xhline} 
\makeatother

\section{Appendix A: Table of Homographs}
\label{section:AppendixHomographTable}
We present three tables of homographs, corresponding to the three categories of homographs discussed within the paper (segmentation ambiguity, morphosyntactic ambiguity, and semantic ambiguity).\footnote{It should be noted that the higher level of ambiguity are supersets of the lower levels: segmentation ambiguities generally entail differences on the morphosyntactic and semantic levels as well, and morphosyntactic ambiguities generally also entail semantic ambiguities. Furthermore, because many of the homographs admit to more than two analyses, it is often the case that a subset of the analyses may form a lower level of ambiguity (e.g., just a semantic ambiguity), while other analyses form a higher level of ambiguity (e.g. segmentation ambiguity). For the purposes of this paper, we categorize each homograph according to the highest level of ambiguity involved: First, if a segmentation ambiguity is indicated anywhere across the possible analyses, then we include the homograph in the "segmentation ambiguity" category. Next, if there is no segmentation ambiguity, but a morphosyntactic ambiguity is indicated anywhere across the possible analyses, then we include the homograph in the "morphosyntactic ambiguity" category. Finally, if the analyses all differ only on the semantic level, then we include the homograph in the "semantic ambiguity" category.} In each table, the first column ("form") indicates the homograph as it is found in naturally-occurring non-diacritized Hebrew text.\footnote{The ranking of analyses is based on a frequency analysis of our in-house annotated corpus. It is worth emphasizing that the paper as a whole relates to each of the 75 homographs specifically as they are spelled in this list, and does not relate to cases where further prefixes are attached to the homographs. As a result, the frequency analysis may sometimes seem counterintuitive. For instance, regarding the form \<r'/sy>, a native Hebrew speaker might intuit that the adjectival form is primary (\<rA'+siy>). However, in practice, that sense is common only when prefixed with a definitive marker (\<hr'/sy>). In contrast, the homograph considered here involves the form  \<r'/sy> as is, without any prefixes; in this case, the other analyses are far more common.} The second column ("word") indicates the possible diacritizations of each form. In cases where the diacritized word includes an attached prefix, a plus sign indicates the segmentation point between the prefix letters and the primary word. Regarding all homographs considered in this paper, different segmentation options are diacritized differently. Thus, for each sentence in the dataset, our human annotators were asked simply to choose the correct diacritization for the target homograph (that is, they were asked to choose among the options listed in the "word" column). There was no need for the annotators to separately tag the segmentation, because in all cases the choice of diacritization itself indicates the segmentation.\footnote{It should be noted that in certain sentences, an exceedingly rare diacritization was warranted, which was not among the options listed in the "word" column. The annotators were instructed to tag such cases as "none of the above", and all such sentences were removed from the corpus. Similarly, some sentences do not provide enough context to determine the correct diacritization; the annotators were asked to tag such sentences as "unclear", and these sentences too were removed from the corpus.
} The third column indicates the morphology of each of the possible diacritizations.\footnote{In most cases, a diacritized form has one specific morphological analysis. However, in other cases, the diacritized form can admit to multiple morphologies. In such cases, we list all of the practically relevant morphological analyses in the third column, separated by a slash (thus for instance in the case of \<zr>). Rare analyses which hardly ever occur in practice are not listed.} The fourth column lists the translation.\footnote{Naturally, a given Hebrew term often captures a substantial range of potential English translations, and it would not be practical to list them all in this column; therefore, we generally present only a single representative translation.} Within each table the homographs are listed alphabetically.\footnote{Ideally, we might have grouped the homographs based on part-of-speech instead. However, as can be seen from the following tables, the 75 homographs vary so widely in terms of the parts of speech that they can represent, such that alphabetical listing was deemed most useful.}

\subsection{Homographs with Segmentation Ambiguity}
\begin{longtable}{|c"c|c|c|c|}\hline
Form & Word & Morphology\footnote{Key to morphology abbreviations: F=Feminine; M=Masculine; S=Singular; P=Plural; abs=absolute; cons=construct; 1,2,3=First Person, Second Person, Third Person; suf=suffix; Det = determiner; SConj=Subordinating Conjunction; Adj=Adjective; Conj=Conjunction; Prep=Preposition; Propn=Proper Noun; Adv=Adverb; Ptcp=Participle; ACC=Accusative marker.} & Translation & \#Sentences \\ \hline\hline
\<h'M>
 & \<ha'iM> & Interrogative & does & 1,000\\ \cline{2-5}
 & \<'eM>+\<hA> & Det + Noun [F,S,abs] & the + mother & 1,000\\  \hline\hline
\<hmr'h>
 & \<m*ar:'Eh>+\<ha> & \small SConj + Ptcp [M,S] / Det + Noun [M,S,abs] & that + indicates / the + sight & 1,000\\ \cline{2-5}
 & \<m*ar:'Ah>+\<ha> & \small SConj + Ptcp [F,S] / Det + Noun [F,S,abs] & 
 that + indicates / the + mirror & 882\\ \cline{2-5}
 & \<ham:rA'Ah> & Noun [F,S,abs] & takeoff & 735\\ 
 \hline
 \newpage
 \hline
 \<hqph>
 & \<q*ApEh>+\<ha> & Det + Noun [M,S,abs] & the + coffee & 1,000\\ \cline{2-5}
 & \<haq*ApAh> & Noun [F,S,abs] & credit & 500\\ \hline\hline
\<hq/sr>
 & \<q*E+sEr>+\<ha> & Det + Noun [M,S,abs] & the + connection & 1,000\\ \cline{2-5}
 & \<q*a+s*Ar>+\<ha> & Det + Noun [M,S,abs] & the + signaler & 1,000\\ \cline{2-5}
 & \<hEq:+ser> & Noun [M,S,abs/cons] & context & 588\\ \hline\hline
\<h/slmh>
 & \<ha+s:lAmAh> & Noun [F,S,abs] & completion & 1,000\\ \cline{2-5}
 & \<+s*:lemAh>+\<ha> & Det + Adj [F,S] & the + complete & 906\\ \hline\hline
\<w`d>
 & \<`ad>+\<w:> & Conj + Prep & and + until & 1,000\\ \cline{2-5}
 & \<wa`ad> & Noun [M,S,abs/cons] & committee & 519\\ \hline\hline
\<l/sM>
 & \<l:+seM> & Prep &  for the purpose of & 1,000\\ \cline{2-5}
 & \<+sAM>+\<l:> & Prep + Adverb &  to + there & 1,000\\ \hline\hline
\<mb.hynh>
 & \<b*:.hiynAh>+\<mi> & Prep + Noun [F,S,abs] & from + point of view & 1,000\\ \cline{2-5}
 & \<mab:.hiynAh> & Participle [F,S,abs] & she notices & 607\\ \hline\hline
\<msybwt>
 & \<m:siyb*wot> & Noun [F,P,abs] & parties & 1,000\\ \cline{2-5}
 & \<s*iyb*wot>+\<mi> & Prep + Noun [F,P,abs/cons] & due to + reasons & 1,000\\ \hline\hline
\<mpt.h>
 & \<map:t*e.ha> & Noun [M,S,abs] & key & 1,000\\ \cline{2-5}
 & \<m:pat*e.ha> & Participle [M,S,abs] & develops / developer & 329\\ \cline{2-5}
 & \<p*Eta.h>+\<mi> & Prep + Noun [M,S,abs/cons] & from + opening & 206\\ \hline\hline
\</s'lh>
 & \<+s:'elAh> & Noun [F,S,abs] & question & 1,000\\ \cline{2-5}
 & \<+sA'a:lAh> & Verb [F,S,3,Past] & she asked & 1,000\\ \cline{2-5}
 & \<'el*Eh>+\<+sE> & SConj + Pronoun [MF,P,3] & that + these & 228\\ \hline\hline
\</s'P>
 & \<'aP>+\<+sE> & SConj + CConj & for + even & 1,000\\ \cline{2-5}
 & \<+sA'aP> & Verb [M,S,3,Past] & he aspired & 665\\ 
 \hline
 \hline
\</sbh>
 & \<b*Ah*>+\<+sE> & Sconj + Prep [suf=F,S,3] & that + in it & 1,000\\ \cline{2-5}
 & \<+sAbAh> & Verb [F,S,Present/Past] & she returns / she returned & 1,000\\ \hline\hline
\</smN>
 & \<+sEmEN> & Noun [M,S,abs/cons] & oil & 1,000\\ \cline{2-5}
 & \<+sAmeN> & Adj [M,S,abs] & wide & 335\\ \cline{2-5}
 & \<m*iN>+\<+sE> & SConj + Prep & that + from & 207\\ \cline{2-5}
 & \<+s:mAN> & Noun [M,S,abs,suf=F,P,3] & their name & 149\\ \hline\hline
\</smr>
 & \<+sAmar> & Verb [M,S,3,Past] & he guarded & 1,000\\ \cline{2-5}
 & \<+sEmEr> & Propn & Shemer & 872\\ \cline{2-5}
 & \<m*ar>+\<+sE> & SConj + Titular [M,S] & that + Mr. & 224\\ \hline
\end{longtable}
\FloatBarrier
\phantom{This text will be invisible} \\

\subsection{Homographs with Morphosyntactic Ambiguity}
\begin{longtable}{|c"c|c|c|c|}\hline
\centering
Form & Word & Morphology & Translation & \#Sentences \\ \hline\hline
\<'hbh>
 & \<'aha:bAh> & Noun [F,S,abs] & love & 1,000\\ \cline{2-5}
 & \<'Aha:bAh> & Verb [F,S,3,Past] & she loved & 1,000\\ \hline\hline
\<'wkl>
 & \<'owkEl> & Noun [M,S,abs/cons] & food & 748\\ \cline{2-5}
  & \<'w*kal> & Modal [MF,S,1,Future] & I can & 729\\
  \cline{2-5}
  & \<'wokel> & Participle [M,S,abs/cons] & eats & 640\\
 \hline\hline
\<'.hdwt>
 & \<'a:.hAdwot> & Det [F,P,abs] & several & 1,000\\ \cline{2-5}
 & \<'a.h:dw*t> & Noun [F,S,abs] & unity & 1,000\\ 
 \hline
 \newpage
 \hline
\<'.hyw>
 & \<'E.hAyw> & Noun [MF,P,abs,suf=M,S,3] & his brothers & 1,000\\ \cline{2-5}
 & \<'A.hiyw> & Noun [MF,S,abs,suf=M,S,3] & his brother & 774\\ \hline\hline
\<'lymwt>
 & \<'al*iymwot> & Adj [F,P] & violent & 1,000\\ \cline{2-5}
 & \<'al*iymw*t> & Noun [F,S,abs/cons] & violence & 1,000\\ \hline\hline
\<'M>
 & \<'iM> & Conj & if & 1,000\\ \cline{2-5}
 & \<'eM> & Noun [F,S,abs/cons] & mother & 1,000\\ \hline\hline
\<'m.s`y>
 & \<'Em:.sA`ey> & Noun [M,P,cons] & centers of / methods of & 1,000\\ \cline{2-5}
 & \<'Em:.sA`iy> & Noun [M,S,abs] / Adj [M,S] & method / central & 969\\ \hline\hline
\<'mrh>
 & \<'Am:rAh> & Verb [F,S,3,Past] & she said & 1,000\\ \cline{2-5}
 & \<'im:rAh> & Noun [F,S,abs] & a saying & 343\\ \hline\hline
\<'p/sr>
 & \<'Ep:+sAr> & Modal / Adv & possible & 1,000\\ \cline{2-5}
 & \<'ip:+ser> & Verb [M,S,3,Past] & he allowed & 511\\ \hline\hline
\<'t>
 & \<'Et> & ACC & accusative & 1,000\\ \cline{2-5}
 & \<'at*:> & Pronoun [F,S,2] & you & 1,000\\ \hline\hline
\<bhm/sK>
 & \<hEm:+seK:>+\<b*:> & Prep + Noun [M,S,cons] & in + continuation of & 1,000\\ \cline{2-5}
 & \<hEm:+seK:>+\<b*a> & Prep [with Det] + Noun [M,S,abs] & in the + future & 930\\ \hline\hline
\<b.hyy>
 & \<.hay*ey>+\<b*:> & Prep + Noun [M,P,cons] & in + lives of & 1,000\\ \cline{2-5}
 & \<.hay*ay>+\<b*:> & Prep + Noun [M,P,abs,suf=MF,S,1] & in + my life & 1,000\\ \hline\hline
\<b`wlM>
& \<`wolAM>+\<b*A> & Prep [with Det] + Noun [M,S,abs] & in the + world & 1,000\\ \cline{2-5}
 & \<`wolaM>+\<b*:> & Prep + Noun [M,S,cons] & in + a world of & 913\\ \cline{2-5}
  & \<`wolAM>+\<b*:> & Prep + Noun [M,S,abs] &  in + a world & 485\\ \hline\hline
\<bqrb>
 & \<qErEb>+\<b*:> & Prep + Noun [M,S,cons] & in + midst of  & 1,000\\ \cline{2-5}
 & \<q*:rAb>+\<b*a> & Prep [with Det] + Noun [M,S,abs] & in the + battle & 1,000\\ \cline{2-5}
 & \<q:rab>+\<b*i> & Prep + Verb [Bare Infinitive] & in + approaching of & 734\\ \cline{2-5}
 & \<q:rAb>+\<b*i> & Prep + Noun [M,S,abs] & in + a battle & 256\\ \hline\hline
\<gylw>
 & \<g*iyl*w*> & Verb [MF,P,3,Past] & they discovered & 1,000\\ \cline{2-5}
 & \<g*iylwo> & Noun [M,S,abs,suf=M,S,3] & his age & 859\\ \hline\hline
\<dy>
 & \<d*ay> & Adverb & sufficiently & 1,000\\
 \cline{2-5}
 & \<d*ey> & Det [cons] & enough of & 828\\  
 \cline{2-5}
 & \<d*iy>  & Prefix & di- & 808\\ 
\hline\hline
\<hzqN>
 & \<z*AqAN>+\<ha> & Det + Noun [M,S,abs] & the + beard & 739\\ \cline{2-5}
 & \<z*AqeN>+\<ha> & \small Det + Adj [M,S] / Det + Noun [M,S,abs] & \small the + old / the + elderly man & 533\\ \hline\hline
\<h.hl>
 & \<he.hel> & Verb [M,S,3,Past] & he began & 1,000\\ \cline{2-5}
 & \<hA.hel> & Verb [Bare Infinitive] & starting (from) & 1,000\\ \hline\hline
\<hm/snh>
 & \<m*i+s:nEh>+\<ha> & Det + Noun [M,S,abs] & the + deputy & 1,000\\ \cline{2-5}
 & \<m*i+s:nAh>+\<ha> & Det + Propn [F,S,abs] & the + Mishna & 1,000\\ \hline\hline
\<hn.hh>
 & \<han*A.hAh> & Noun [F,S,abs] & placing & 1,000\\ \cline{2-5}
 & \<hin:.hAh> & Verb [M,S,3,Past] & he directed & 731\\ \cline{2-5}
 & \<ha:nA.hAh> & Noun [F,S,abs] & discount & 517\\ \hline\hline
\<hryM>
 & \<heriyM> & Verb [M,S,3,Past] & he lifted & 1,000\\ \cline{2-5}
 & \<hAriyM> & Noun [M,P,abs] & mountains & 1,000\\ \hline\hline
\<w't>
 & \<'Et>+\<w:> & Conj + ACC & and + accusative & 1,000\\ \cline{2-5}
 & \<'at*:>+\<w:> & Conj + Pronoun [F,S,2] & and + you & 1,000\\ \hline\hline
\<zr>
 & \<zer> & Noun [M,S,abs/cons] & bouquet & 1,000\\ \cline{2-5}
 & \<zAr> & Adj [M,S] / Noun [M,S,abs] & foreign / stranger & 1,000\\ 
 \hline
 \newpage
 \hline
\<.hbrwt>
 & \<.ha:bArwot> & Noun [F,P,abs] & companies & 1,000\\ \cline{2-5}
 & \<.hEb:rwot> & Noun [F,P,cons] & companies of & 1,000\\ \cline{2-5}
 & \<.ha:berwot> & Noun [F,P,abs/cons] & friends & 501\\ \cline{2-5}
 & \<.ha:berw*t> & Noun [F,S,abs/cons] & friendship & 398\\ \cline{2-5}
 & \<.hab:rwot> & Noun [F,P,cons] & friends of & 261\\ \hline\hline
\<.hdr>
 & \<.ha:dar> & Noun [M,S,cons] & room of & 1,000\\ \cline{2-5}
 & \<.hEdEr> & Noun [M,S,abs] & room & 1,000\\ \cline{2-5}
 & \<.hAdar> & Verb [M,S,3,Past] & penetrated & 783\\ \hline\hline
\<.twb>
 & \<.twob> & Adj [M,S] & good & 1,000\\ \cline{2-5}
 & \<.tw*b> & Noun [M,S,abs/cons] & goodness & 357\\ \hline\hline
\<yhwdy>
 & \<y:hw*diy> & Noun [M,S,abs] / Adj [M,S] & a Jew / Jewish & 1,000\\ \cline{2-5}
 & \<y:hw*dey> & Noun [M,P,cons] & Jews & 1,000\\ \hline\hline
\<kywwN>
  & \<k*eywAwN> & Conj & because & 487\\ 
 \cline{2-5}
& \<k*iyw*w*N> & Noun [M,S,abs] / Noun [M,S,cons] & direction & 468\\
  \cline{2-5}
& \<k*iyw*ewN> & Verb [M,S,3,Past] & directed & 455\\  \hline\hline
\<lw>
 & \<lwo> & Prep [suf=M,S,3] & to him & 1,000\\ \cline{2-5}
 & \<lw*> & Conj & if only & 1,000\\ \hline\hline
\<l.hM>
 & \<lE.hEM> & Noun [M,S,abs] & bread & 1,000\\ \cline{2-5}
 & \<lA.haM> & Verb [M,S,3,Past] & he fought & 1,000\\ \hline\hline
\<lpnwt>
 & \<lip:nwot> & Prep / Verb [Infinitive] & facing / to turn & 1,000\\ \cline{2-5}
 & \<l:pan*wot> & Verb [Infinitive] & to clear out & 570\\ \hline\hline
\<mdy>
 & \<mid*ey> & Det [cons] & every & 1,000\\ \cline{2-5}
 & \<mid*ay> & Adv & too much & 802\\ \cline{2-5}
 & \<mad*ey> & Noun [M,P,cons] & uniforms of & 541\\ \hline\hline
\<mhM>
 & \<mehEM> & Preposition [suf=M,P,3] & from them & 1,000\\ \cline{2-5}
 & \<mAheM> & Interrogative & what are & 587\\ \hline\hline
\<my>
 & \<miy> & Interrogative / Pronoun [S,3] & who & 1,000\\ \cline{2-5}
 & \<mey> & Noun [M,P,cons] & waters of & 1,000\\ \hline\hline
\<mlK>
 & \<mElEK:> & Noun [M,S,abs/cons] & king & 1,000\\ \cline{2-5}
 & \<mAlaK:> & Verb [M,S,3,Past] & he ruled & 522\\ \hline\hline
\<m`br>
 & \<me`ebEr> & Prep & beyond & 1,000\\ \cline{2-5}
  & \<ma`a:bAr> & Noun [M,S,abs] & passage & 1,000\\ 
  \cline{2-5}
  & \<ma`a:bar> & Noun [M,S,cons] & passage of & 883\\ 
\hline\hline
\<mr'h>
 & \<mar:'Eh> & Participle [M,S] & he shows & 1,000\\ \cline{2-5}
 & \<mar:'Ah> & Participle [F,S] & she shows & 1,000\\ \hline\hline
\<mrkz>
 & \<mEr:k*az> & Noun [M,S,cons] & center of & 1,000\\ \cline{2-5}
 & \<mEr:k*Az> & Noun [M,S,abs] & center & 1,000\\ \cline{2-5}
 & \<m:rak*ez> & Participle [M,S,abs/cons] & organizes / organizer & 393\\ \hline\hline
\<m/s.hq>
 & \<mi,s:.hAq> & Noun [M,S,abs] & game & 1,000\\ \cline{2-5}
 & \<m:,sa.heq> & Participle [M,S,abs] & plays / player & 479\\ \hline\hline
\<n`/sh>
 & \<na`a:,sEh> & Verb [MF,P,1,Future] & we will do & 1,000\\ \cline{2-5}
 & \<na`a:,sAh> & Verb [M,S,3,Past] & was done & 1,000\\ \hline\hline
\<n/syM>
 & \<nA+siyM> & Noun [F,P,abs] & women & 1,000\\ \cline{2-5}
 & \<nA,siyM> & Verb [MF,P,1,Future] & we will put & 613\\ \hline\hline
\<ntN>
 & \<nAtaN> & Verb [M,S,3,Past] & gave & 1,000\\ \cline{2-5}
 & \<nAtAN> & Propn & Nathan & 681\\ 
 \hline
 \newpage
 \hline
\<`br>
 & \<`Abar> & Verb [M,S,3,Past] & he passed & 1,000\\ \cline{2-5}
 & \<`AbAr> & Noun [M,S,abs] & past & 1,000\\ \cline{2-5}
 & \<`ebEr> & Noun [M,S,abs/cons] & side & 456\\ \hline\hline
\<`d>
  & \<`ad> & Prep & until & 1,000\\ 
\cline{2-5}
& \<`ed> & Noun [M,S,abs/cons] & witness & 1,000\\  \hline
 \hline
\<`wbdwt>
 & \<`uwb:d*wot> & Noun [F,P,abs/cons] & facts & 1,000\\ \cline{2-5}
 & \<`wob:dwot> & Participle [F,P] & they work / workers & 1,000\\ \hline\hline
\<`M>
 & \<`iM> & Prep & with & 1,000\\ \cline{2-5}
 & \<`aM> & Noun [M,S,abs/cons] & nation & 1,000\\ \hline\hline
\<pny>
 & \<p*:ney> & Noun [M,P,cons] & faces of & 1,000\\ \cline{2-5}
 & \<p*Anay> & Noun [MF,P,abs,suf=MF,S,1] & my face & 338\\ \hline\hline
\<prs>
 & \<p*:rAs> & Noun [M,S,abs] & award & 1,000\\ \cline{2-5}
 & \<p*ErEs> & Propn & Peres & 956\\ \cline{2-5}
 & \<p*Aras> & Verb [M,S,3,Past] & he spread & 630\\ \cline{2-5}
 & \<p*:ras> & Noun [M,S,cons] & award of & 290\\ \hline\hline
\<.sywN>
 & \<.siy*woN> & Propn & Zion & 1,000\\ \cline{2-5}
 & \<.siy*w*N> & Noun [M,S,abs/cons] & mark & 1,000\\ \hline\hline
\<qwdM>
 & \<qowdEM> & Adv & before & 1,000\\ \cline{2-5}
 & \<qwodeM> & Adj [M,S] & previous & 1,000\\ \cline{2-5}
 & \<quwd*aM> & Verb [M,S,3,Past] & was promoted & 284\\ \hline\hline
\<r'/sy>
 & \<rA'+sey> & Noun [M,P,cons] & heads & 1,000\\ \cline{2-5}
 & \<ro'+siy> & Noun [M,S,abs,suf=MF,S,1] & my head & 881\\ \cline{2-5}
 & \<rA'+siy> & Adj [M,S,abs] & head & 399\\ \hline\hline
\</syrt>
 & \<+seyret> & Verb [M,S,3,Past] & he served & 1,000\\ \cline{2-5}
 & \<+siyrat> & Noun [F,S,cons] & poetry of & 896\\ \hline\hline
\</skr>
 & \<,sAkAr> & Noun [M,S,abs] & salary & 1,000\\ \cline{2-5}
 & \<,s:kar> & Noun [M,S,cons] & salary of & 751\\
 \cline{2-5}
   & \<,sAkar> & Verb [M,S,3,Past] & rented & 681\\\hline\hline
\</sM>
 & \<+seM> & Noun [M,S,abs/cons] & name & 1,000\\ \cline{2-5}
 & \<+sAM> & Adv & there & 1,000\\ \cline{2-5}
 & \<,sAM> & Verb [M,S,Present/Past] & he placed & 1,000\\ \hline\hline
\<tn'y>
 & \<t*:nA'ey> & Noun [M,P,cons] & conditions of & 1,000\\ \cline{2-5}
 & \<t*:na'y> & Noun [M,S,abs/cons] & condition  & 834\\ \hline
\end{longtable}

\FloatBarrier
\phantom{This text will be invisible} \\

\subsection{Homographs with Semantic Ambiguity}
\begin{longtable}{|c"c|c|c|c|}\hline
Form & Word & Morphology & Translation & \#Sentences \\ \hline\hline
\<hzmr>
 & \<z*EmEr>+\<ha> & Det + Noun [M,S,abs] & the + music & 1,000\\ \cline{2-5}
 & \<z*am*Ar>+\<ha> & Det + Noun [M,S,abs] & the + musician & 1,000\\ 
 \hline
 \newpage
 \hline
\<hswpr>
 & \<s*w*p*Er>+\<ha> & Det + Noun [M,S,abs] & the + market & 763\\ \cline{2-5}
 & \<s*woper>+\<ha> & Det + Noun [M,S,abs] & the + author & 570\\ \hline\hline
\<zmr>
 & \<zam*Ar> & Noun [M,S,abs] & musician & 1,000\\ \cline{2-5}
 & \<zEmEr> & Noun [M,S,abs] & song & 602\\ \hline\hline
\<.hbrh>
 & \<.ha:berAh> & Noun [F,S,abs] & friend & 1,000\\ \cline{2-5}
 & \<.hEb:rAh> & Noun [F,S,abs] & company & 1,000\\ \hline\hline
\<r/swt>
 & \<r:+sw*t> & Noun [F,S,abs/cons] & permission & 1,000\\ \cline{2-5}
 & \<rA+sw*t> & Noun [F,S,abs/cons] & authority & 1,000\\ \hline
\end{longtable}

\FloatBarrier
\phantom{This text will be invisible} \\


\twocolumn
\section{Appendix B: Creation of our Annotated Dataset}
\label{section:appendixCuration}
We present here further details regarding the creation of our annotated homograph dataset. As noted above, our dataset contains challenge sets for 75 Hebrew homographs. In almost all cases, we collect 1000 sentences for the primary analysis, at least 500 sentences for the secondary analysis, and at least 250 for each additional analysis. Every sentence was tagged by our team of human annotators.

The main challenge that we faced is that it is often extraordinarily difficult to identify a sufficient number of naturally-occurring Hebrew sentences attesting to the non-primary homograph analyses. In many cases, the naturally-occurring ratio of these analyses is 100:1 or worse, meaning that the non-primary analyses only occur once in 100 or more sentences. In such cases, prima facie, we would need to tag some 50,000 randomly-selected cases of the homograph in order to gather 500 cases of the alternate analysis. It would not be practical to tag this number of sentences for each of the 75 homographs. Therefore, we devised the following three-step process to allow us to efficiently gather a sufficient number of sentences attesting each possible analysis of each homograph:

1. For each homograph, we first gather 4000 randomly-selected sentences from an untagged Hebrew corpus. For each sentence, our annotation team determines the correct analysis of the homograph. This initial step virtually always provides us with a sufficient number of sentences attesting to the primary analysis, and sometimes for the secondary analysis as well. However, it often leaves us with an insufficient number of sentences regarding the non-primary analyses. 

2. In order to find additional sentences for the non-primary homograph analyses, we train classifiers to identify a wide net of potential candidate sentences. For these classifiers, we use the known cases of the primary homograph analysis (identified in the previous step) as the first class, and we use various proxies to represent the opposing class. The proxies used for the opposing class include the following: (A) Hebrew words which unambiguously function morphosyntactically in a manner that is different from the morphosyntactic function of the primary class; (B) Hebrew words whose word2vec embedding is maximally distant from that of the homograph in question (because, in practice, for homographs of skewed proportions, the word2vec embedding of the word tends to represent the primary usage); (C) Randomly-selected Hebrew words. These classifiers are trained with a BiLSTM encoding of the word2vec embeddings of the four neighbors on either side of the target word (not including the target word itself). Essentially, the point of these classifiers is to identify a generous selection of cases in which the homographs may possibly be functioning in a way that is different than their primary usage.

3. We run these classifiers on all instances of the target homograph in our untagged Hebrew corpus. Cases that are classified as the opposing class are sent to our annotators for human verification. In practice, these proxy-based classifiers reach a precision of 15-30\% on the task of identifying sentences in which the target homographs function according to a non-primary analysis. Thus, on average, for any given homograph, tagging an additional 2222 sentences allows us to collect 500 relevant cases of the secondary analysis. 

Our procedure ensures diversity while keeping the process cost-effective.  All in all it took approximately 1500 hours of annotation time to amass the sentences in the present dataset.

\section{Appendix C: Computational Equipment}
We performed all computations on a desktop workstation with an i9-10980XE processor and 256GB of memory. This system enabled us to run 36  experiments in parallel (the processor contains 18 hyperthreaded cores), and thus we were able to complete all of the relevant experiments and computations over the course of several weeks of calendar time. 

\section{Appendix D: Neural Network Implementation Details}
\label{section:appendixImplementation}
All BiLSTMs and MLPs were trained using dynet (http://dynet.io/). All MLPs are constructed with a hidden layer of size 100, and with two layers. We train with the Adam optimizer at a learning rate of .001, for 3 epochs. 

Our word2vec embeddings for Hebrew are of size 100. We trained them on a 500M-word Hebrew corpus using word2vecf (\url{https://github.com/BIU-NLP/word2vecf}, adding position info to context words, and excluding words with a frequency of less than seven. All in all, our word2vec vocabulary covers 94.6\% of the tokens in our dataset. For the out-of-vocabulary words, we use a trainable UNK parameter in place of the word2vec embedding, which is trained from scratch for each “word expert” classifier.

As per the "word expert" paradigm, a completely separate MLP is trained for each homograph. In each case, the possible homograph analyses are each treated as a possible class for prediction, and the MLP is trained to choose from among those classes. Thus, for instance, if the homograph has two analyses, we train an MLP to predict one of the two classes; if the homograph has three analyses, then we train an MLP to predict one of the three classes; and so on.

For the Probing Scenarios based on centroid classification, we proceed as follows. For each of the homographs, given the training sample size (100, 50, 25, etc.), we randomly select that amount of training sentences for each of the possible analyses of the homograph. We calculate the centroid for each of the analyses by averaging the embeddings of the target homographs within the corresponding training sentences. The remainder of the available sentences for the homograph forms the test set. We classify them by calculating the dot product of the embedding of the target homograph in each given test sentence with the centroid of each of the homograph analyses. We run this process 200 times, each time selecting a different random set of training sentences. The values plotted in Figure \ref{fig:few-shot} reflect the average of the F1 scores across these 200 rounds. For the corresponding MLP-based experiments presented for comparison in the aforementioned table, we follow an analogous procedure, across 10 rounds.

\end{document}